\documentclass[11pt,a4paper]{article}
\usepackage[hyperref]{acl2018}
\usepackage{times}
\usepackage{latexsym}
\usepackage{graphicx}
\usepackage{url}
\usepackage[english]{algorithm2e}
\usepackage{booktabs}
\usepackage{enumitem}
\graphicspath{ {image/} }
\usepackage{geometry}    
\usepackage{multicol}    
\usepackage{float}   

\aclfinalcopy 

\title{Two Step Joint Model for Drug Drug Interaction Extraction}

\author{
    {Siliang Tang*, Qi Zhang*, Tianpeng Zheng*, Mengdi Zhou*, Zhan Chen**, Lixing}\\
    {Shen***, Xiang Ren***, Yueting Zhuang*, Shiliang Pu** and Fei Wu*}\\
    {*Zhejiang University}\\
    \{siliang, zhangqihit, 21721120, 21721125, yzhuang, wufei\}@zju.edu.cn\\
    {**Hikvision}\\
    \{chenzhan, shenlixing, pushiliang\}@hikvision.com\\
    {***University of Southern California}\\
    xiangren@usc.edu
}

\date{}
\begin{document}
\maketitle

\section*{Abstract}
   When patients need to take medicine, particularly taking more than one kind of drug simultaneously,  they should be alarmed that there possibly exists drug-drug interaction. Interaction between drugs may have a negative impact on patients or even cause death. Generally, drugs that conflict with a specific drug (or label drug) are usually described in its drug label or package insert. Since more and more new drug products come into the market, it is difficult to collect such information by manual. We take part in the Drug-Drug Interaction~(DDI) Extraction from Drug Labels challenge of Text Analysis Conference~(TAC) 2018, choosing task1 and task2 to automatically extract DDI related mentions and DDI relations respectively. Instead of regarding task1 as named entity recognition~(NER) task and regarding task2 as relation extraction~(RE) task then solving it in a pipeline, we propose a two step joint model to detect DDI and it's related mentions jointly. A sequence tagging system~(CNN-GRU encoder-decoder) finds precipitants first and search its fine-grained Trigger and determine the DDI for each precipitant in the second step. Moreover, a rule based model is built to determine the sub-type for pharmacokinetic interation. Our system achieved best result in both task1 and task2. F-measure reaches 0.46 in task1 and 0.40 in task2.

\section{Introduction}

  A drug interaction is a situation in which a substance (usually another drug) affects the activity of a drug when both are administered together. Drug-drug interactions can lead to a variety of adverse events, and it has been suggested that preventable adverse events are the eighth leading cause of death in the United States. Therefore, prescription made up for patients should be cautious. Usually, precipitants with corresponding interaction are listed in package insert for specific label drugs. However, researchers are difficult to collect total DDI information from biomedical literature manually since numerous drugs are circulated in market. So methods that can automatically extract information from drug texts should be put forward as soon as possible.
  
  With the development of machine learning, some people have tried to apply ML methods to DDI extraction. DDI Extraction 2011 task, as well as DDI Extraction 2013 task, encouraged people to detect DDIs in literature with information extraction (IE) methods. Some work followed the DDI Extraction 2013 task and obtained better result than the state-of-art in 2013 task.
  
  Different from DDI tasks in 2011 and 2013, Drug-Drug Interaction Extraction from Drug Labels in TAC 2018 was divided into four subtasks. We participated in task1 and task2, which were proposed to extract mentions (precipitant, trigger, specificInteraction) and identify interaction types (pharmacokinetic,  pharmacodynamic, unspecified) in sentences respectively. Tradition method to deal with task1 and task2 is firstly using NER system to find mentions in a sentence, then utilizing RE system to identify interaction of label drug and precipitant. But in these two tasks, we notice that the sentence in one drug label document share the same label drug and the label drug is given. So we propose a two-step joint model which only constructed several NER system with the same architecture, getting rid of traditional RE method to obtain relation types of drug pair. We first train a NER system to extract precipitants in raw text. Then for each precipitant, we train other two NER systems with identical structure to find its fine-grained trigger and specificInteraction if it exists. Fine-grained trigger means trigger can be divided into three kinds: pharmacokinetic trigger (PK-T), pharmacodynamic trigger (PD-T) and unspecified (UN-T). These triggers can indicate the relation types directly.
  
  The rest of this paper is organized as follows: we shortly depict two tasks that resemble current tasks in section 2. We elaborate model architecture and strategies we use in section 3. In section 4, we describe data as well as extra resources utilized in our model. We describe the final submission and result in section 5. In section 6, we draw our conclusion.

\section{Related Tasks}
  Beside DDI extraction task in TAC 2018, similar DDI extraction tasks were proposed in 2011 and 2013. The DDIExtraction 2011 challenge task was relatively simple as it only focused on judging whether if there contains DDI information or not with given drug-drug pairs in a sentence. The best performance was achieved by the team WBI ~\cite{re2011}, system of which combines several kernels and a case-based reasoning (CBR) using a voting approach. In 2013, the challenge task added a subtask of recognition and classification of drug names, and it pursued the classification of each drug-drug interaction according to pre-defined types. The best result in NER was achieved by the WBI team with a conditional random field (CRF) ~\cite{2013wbi}, while best performance in RE was submitted by the team from FBK-irst ~\cite{2013fbk} with hybrid kernel based RE approach. Both of subtasks in 2013 resemble task1 and task2 in this year, but there are some differences in detail. The first subtask in 2013 need to classify drug names by four types (drug, brand, group, drug-n), while current task1 only requests to recognize precipitants. Present task1 seems to simplify the recognition of drugs, but label drugs have to be filtered. In addition, it is necessary to extract triggers and specificInteractions. In the aspect of relation extraction, present task2 is more challenging as its result relies on the accuracy of task1’s result, while the second subtask in 2013 only identified DDI relations based on given drug-drug pairs as drug entities had been annotated. 

\section{System Architecture}

\begin{figure*}[t]
    \centering
    \includegraphics[width=\textwidth]{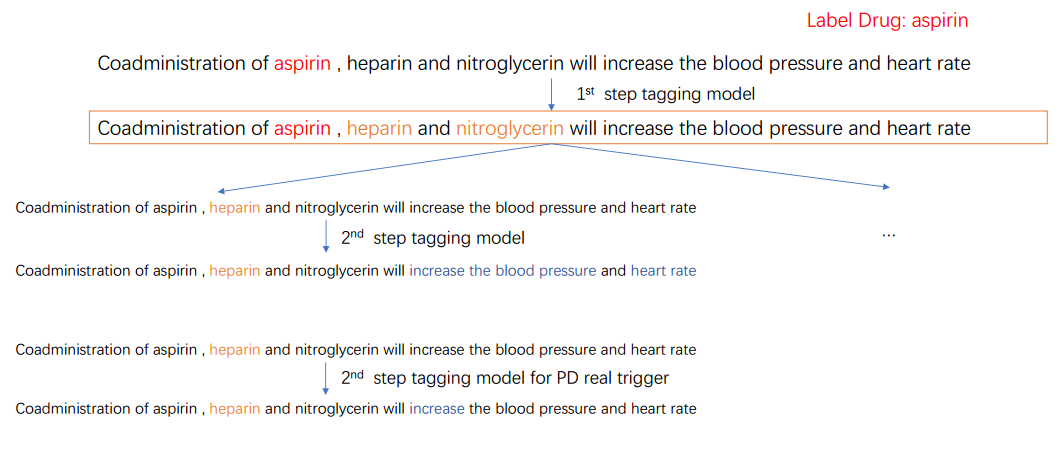}
    \caption{An example of two-step joint model applying to a sentence}
    \label{fig:p1}
\end{figure*}

  We propose a two-step joint model to identify three kinds of mentions defined in Task 1 and three kinds of interactions defined in Task 2. An example is showed in Figure \ref{fig:p1}. We notice that precipitant is the core attribute of each interaction, i.e., each interaction contains its corresponding precipitant which will interact with label drug although the label drug does not exist explicitly in this sentence. Each drug-drug interaction we want to extract has different information from other drug-drug interactions. We try to understand drug-drug interaction extraction task from human perspectives. First, determine whether any drug exists in the given sentence including drug surrogates such as abbreviations and generic/trade name equivalents. Second, determine whether the identified drug is the label drug or not. If it’s not the label drug, then determine whether the trigger of drug-drug interaction exists in the given sentence. If it does, we can extract one drug-drug interaction and get its corresponding attribute.
  
  Two-step joint model consists of two sequence labeling systems. According to the analysis above, the first step of two-step joint model is to find the trigger location based on the semantic information of given sentence and the position features of label drug. The second step is to construct some features by using the precipitant found in the first step and recognize the related trigger. Once the trigger is not mentioned in the sentence, the precipitant found in the first step will be discarded.
  
  The triggers of PK interactions and Unspecified interactions annotated by official standard guideline sufficiently indicate the existences of these two kinds of drug-drug interactions while most of the triggers of PD don’t, for example, ``increase of'' is a trigger of PD interaction but it does not provide enough evidence of a drug-drug interaction. Therefore, we regard the SpecificInteraction attribute of PD interaction as the trigger word in the second step. Another sequence labeling system is applied to find the real trigger attributes of PD interactions as Figure \ref{fig:p1} shows. For one precipitant, its related real trigger and SpecificInteraction will be merged to generate final result.
  
\subsection{Construct Input Features}

\subsubsection{Input Features of First Step}
In the first step, the sequence labeling system utilizes two features. The first feature divides each word into four types, UPPER (all letters of the word are capitalized), UPPER-FIRST (only the first letter of the word is capitalized), LOWER (all the letters of the word are lowercase) and NUM-PUNC (representing the word as a number or a punctuation, e.g. 86\%). This feature can be regarded as an extension of capitalization feature. The second feature is the distance feature between label drug and the current word. The distance is directly defined as the number of words between label drug and the current word. The distance is mapped to the final coarse-grained position features of label drug and it's value is  calculated according to a predefined number of intervals, such as $(5m, 5m+ 5)$ interval maps the value between $5m$ and $5m+5$ to $m$ as the eigenvalue input $(m = 1,2,...)$. When label drug does not exist in the text, the feature value is uniformly set to 200.

\subsubsection{Input Features of Second Step}

The second step of two-step joint model is to recognize the associated trigger based on identified precipitant in the first step, thereby determining the drug-drug interaction type simultaneously. We construct features in each time-step, including word embeddings, coarse-grained position features of precipitants and character embeddings.

\subsection{Tagging Schema}

\begin{figure}[h]
    \centering
    \includegraphics[width=0.50\textwidth]{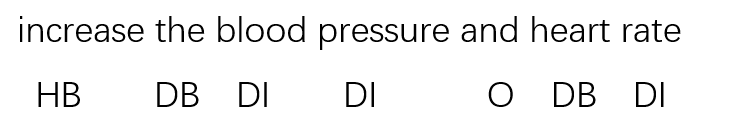}
    \caption{Examples of BIOHD tagging}
    \label{fig:biohd1}
\end{figure}

Commonly, in named entity recognition systems, annotated data is encoded using BIO tagging, where each word is assigned into one of three labels: B means beginning, I means inside, and O means outside of a concept. However, BIO encoding is not sufficient because mentions in DDI task are often disjoint concepts with overlapping words or discontinuous spans. For example, the sentence ``increase the blood pressure and heart rate'' contains two SpecificInteraction entities, ``increase the blood pressure'' and ``increase $|$ heart rate''. The entity phrase ``increase $|$ heart rate'' is a disjoint mention and has a shared word ``increase'' with the other entity phrase ``increase the blood pressure''. To handle those disjoint mentions, we apply an alternative label encoding schema called BIOHD, where H means overlapping portions and D means shared portions. BIOHD encoding label set can be written as $\{H, D\} \times \{B, I\} \bigcup \{O\}$. Figure \ref{fig:biohd1} shows an example of BIOHD label encoding with semantic type and annotated mentions.

\begin{figure}[h]
    \centering
    \includegraphics[width=0.50\textwidth]{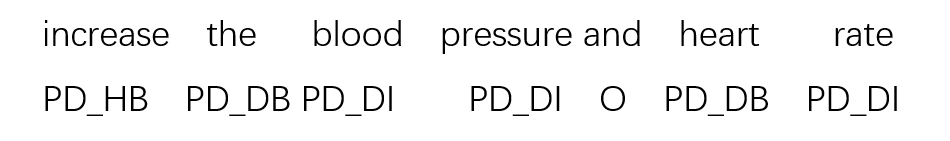}
    \caption{Examples of fine-grained BIOHD tagging}
    \label{fig:biohd2}
\end{figure}

In the second step of two-step joint model, we need to determine the type of the trigger while finding the trigger to determine the corresponding drug-drug interaction type. Now the encoding label set is further expanded to $\{\{H,D\} \times \{B,I \}\bigcup \{O\} \}x \{PK,PD,UN\}$, which is the Cartesian product of labels and DDI categories. Figure \ref{fig:biohd2} shows an example of a sentence with further expanded encoding label set. 

Since there are only three types of drug-drug interactions, the number of labels is not very large even if the interaction type is encoded into a label.

\subsection{Sequence Labeling Model}

\begin{figure}[h]
    \centering
    \includegraphics[width=0.50\textwidth]{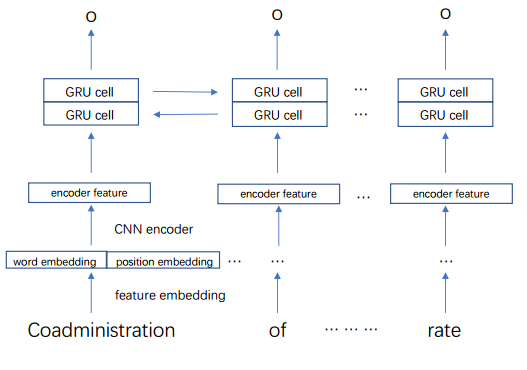}
    \caption{Sequential tagging model}
    \label{fig:p2}
\end{figure}

In the two steps of the two-step joint model, we used the same sequence labeling model to learn the mapping from input features to labeling sequences. The model consists of an encoder based on convolutional neural networks (CNN) and a decoder based on recurrent neural networks (RNN). The CNN decoder passes the  concatenated embedding result of all the features of each time-step through three convolutional layers, and finally generates the coding features of each step through ReLU (Rectified Linear Unit) active layer. The RNN decoder decodes the features generated by the of each step into the label set by using the bidirectional GRU (Gated Recurrent Unit) network. The model is illustrated with Figure \ref{fig:p2}.

\subsection{Hyperparameter Setting}

We tune the model using cross validation on the available dataset and use grid search to determine the optimal parameters. All hyperparameters are detailed in the Table ~\ref{hyper-table}.
\begin{table}[t!]
\begin{center}
\begin{tabular}{|l|l|}
\hline \bf Parameter & \bf Setting \\ \hline
Early stopping number & 10  \\
Decoder output size & 28 \\
Encoder output size & 400 \\
Beam size & 8 \\
Encoder filter size & 3 \\
Dropout rate & 0.25 \\
Batch size & 32 \\
\hline
\end{tabular}
\end{center}
\caption{\label{hyper-table}Parameter Setting}
\end{table}

\begin{figure*}[h]
    \centering
    \includegraphics[width=\textwidth]{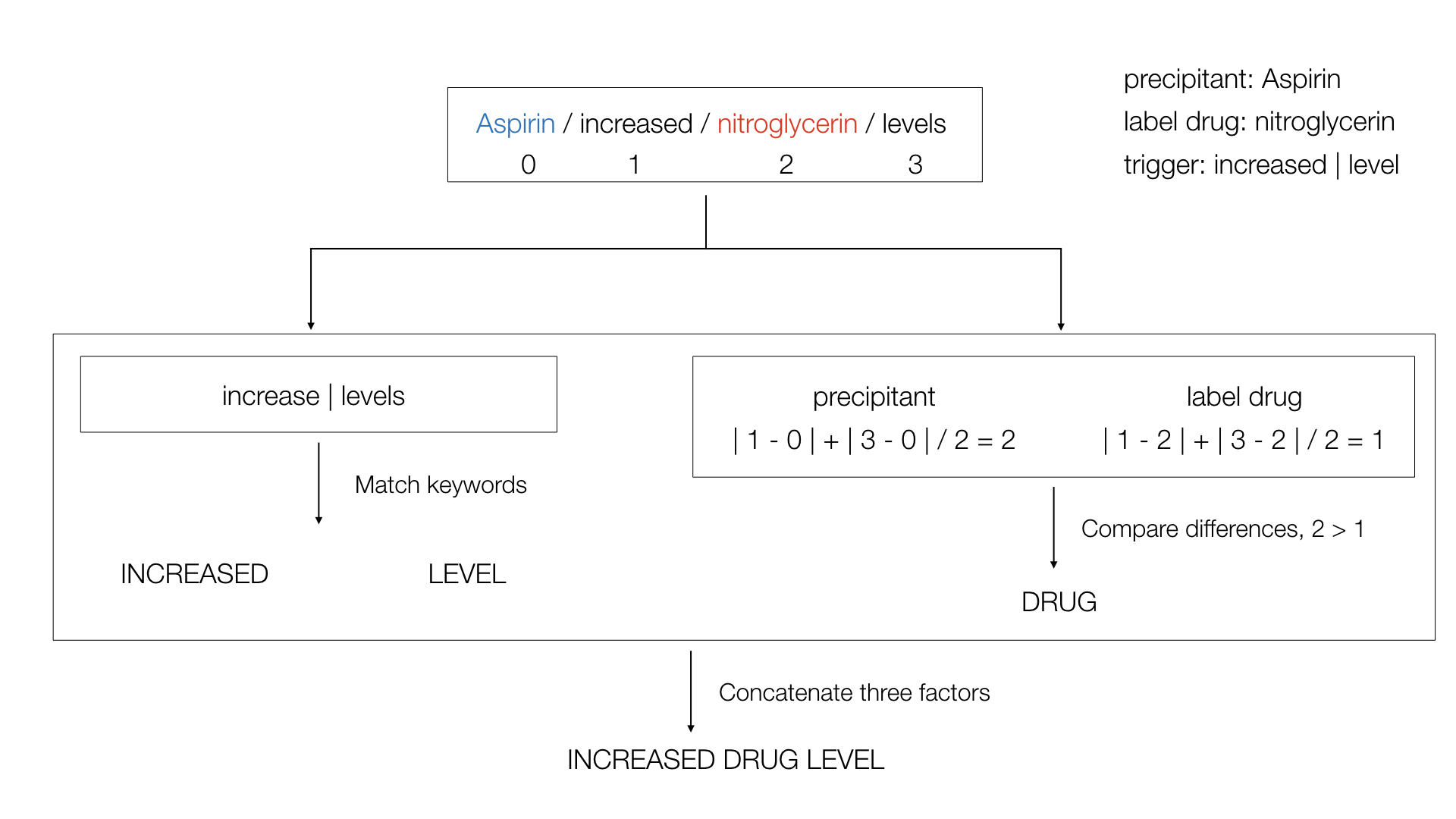}
    \caption{An example of predicting the PK subtype}
    \label{fig:p4}
\end{figure*}

\subsection{PK Subtype Classification}

We design a pharmacokinetic (PK) subtype classification model to determine the subtype of PK because PK interactions must be further classified and coded according to an FDA picklist based on DDI validation guideline. 

PK interactions consist of 20 subtypes depending on three factors, whether the trend is INCREASED or DECREASED, which PK parameter (AUC, Cmax, half-life, LEVEL and Tmax) is affected and whether the effect is on the DRUG or CONCOMITANT DRUG. For a certain PK interaction, its corresponding sentence, trigger, precipitant, label drug which is implicitly or explicitly described in the sentence are fed into PK subtype classification model.

Considering that the trigger word itself contains enough information to predict the former two factors, a dictionary-based method is used to decide whether the trend is INCREASED or DECREASED and which PK parameter is affected. We collect trigger keywords from training dataset and then build up two dictionaries respectively called trend dictionary and parameter dictionary. Then match the triggers with items in two dictionaries and predict the former two factors. For instance, the trigger phrase ``reduces $|$ plasma concentrations'' contains a trend keyword ``reduces'' and a parameter keyword ``concentrations'', which strongly demonstrates that the trend of this PK interaction is DECREASED and the affected parameter “concentrations” belongs to LEVEL according to official guideline.

As for the third factor, we apply a rule-based method, compute distance differences between drugs and trigger to decide whether the effect is on the DRUG or CONCOMITANT DRUG. In general, when the label drug is not mentioned explicitly in the text, we reasonably predict that the effect is on the CONCOMITANT DRUG. Otherwise, compare the difference between label drug and trigger with the difference between concomitant drug (i.e. precipitant) and trigger and find the smaller one. If the drug is label drug or its surrogate, we decide the effect is on the DRUG. If the drug is precipitant, we decide the effect is on the CONCOMITANT DRUG.

Combining the above results of three factors, we can get the subtype for a certain PK interaction and code it according to the FDA picklist. There is an illustration of how we predict PK subtype in Figure \ref{fig:p4}. Table 2 shows some examples.

\begin{table*}
\centering
\begin{tabular}{llll}
  Examples of PK triggers & Trend keyword & Parameter keyword & Prediction\\
  \hline
  increases exposure & increases & exposure & INCREASED LEVEL\\
  elevated plasma concentrations & elevated & concentrations & INCREASED LEVEL\\
  decreases exposure & decreases & exposure & DECREASED LEVEL\\
  lower serum levels & lower & levels & DECREASED LEVEL\\
  increased Cmax & increased & Cmax & INCREASED CMAX \\
\end{tabular}
\caption{Examples of PK triggers matching with the trend dictionary and parameter dictionary.
  }
\end{table*}

\begin{figure*}[h]
    \centering
    \includegraphics[width=\textwidth]{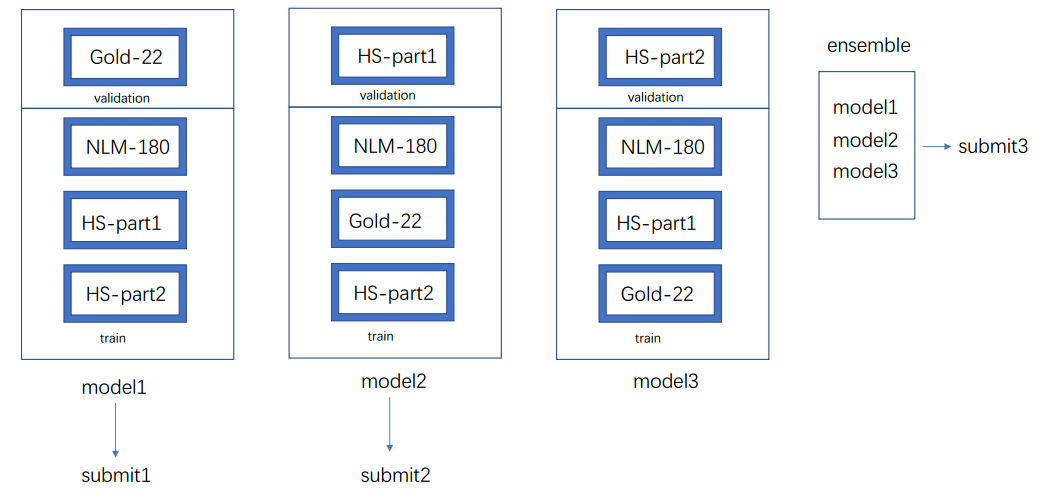}
    \caption{Ensemble method to determine the best sequence}
    \label{fig:p3}
\end{figure*}

\section{Data and Extra Resources}

\subsection{Data}
After careful observation and analysis, there are some problems in Training-22 dataset annotated in gold standard format, including sample duplication, annotation incorrectness and mention annotation inconsistency with official validation guideline, especially trigger annotations. To address the above issues, we manually correct the mention annotations which don’t comply with guideline specifications. In addition, data augmentation is done because the scale of Training-22 dataset is not very large.

We transform NLM-180 dataset into Training-22 format according to certain rules, thus augmenting the dataset. There are some differences between NLM-180 and Training-22 in annotation format. In NLM-180, label drugs are explicitly annotated. On the contrary, in Training-22, label drugs are not labeled explicitly in the extracted drug-drug interactions and are defined at the beginning of each document. Thus, annotated label drugs are filtered out during our transformation from NLM-180 format to Training-22 format. Meanwhile, the definition of trigger in NLM-180 is distinct from that in Training-22, especially the trigger of PD. In fact, the trigger of PD in NLM-180 corresponds to SpecificInteraction of PD in Training-22 and NLM-180 lacks real trigger field corresponding to trigger of PD in Training-22. Therefore, during the transformation we manually add trigger field for PD according to the guideline standard.

The previous related DDI datasets can’t be used as supplementary resources directly due to the unique data format of this competition task. Since the datasets provided by organizers are from DailyMed, we download some free resources from DailyMed and preprocess the raw texts. We annotate the data at sentence level with reference to the official DDI validation guideline. We sample numerous documents but only sample several sentences in one document to guarantee  data diversity and richness. We are not professional in medicine, a highly specialized domain, so plenty of positive samples are possibly omitted and it is inevitable that there are some annotation errors which hardly affect our model nevertheless. The number of sentences we annotate is approximately 1100 and we split it into two equal amount parts annotated as HS-part1 and HS-part2.

\subsection{Extra Resources}
We used some extra resources during training process, including PubMed, PubMed Central (PMC) corpus, Google Word2vec tool to train word vectors.  

\section{Final Submission}
For submission, we divide the available dataset into several parts and train the model with Adadelta algorithm, adjusting the learning rate dynamically. Figure \ref{fig:p3} illustrates dataset partition results. we use the predict result of model1 and model2 as the first and second submission. One of the partitions serves as a validation set to determine the hyper-parameters of the model. Beam search is used as a trade-off of greedy search decoding method and Viterbi decoding method during the test time.

In the third version, we designed a simple sequence decoding ensemble algorithm to integrate the prediction results of the above three models. In this ensemble algorithm, let $S_{1,1}$ be the score of the best sequence $MS_{1}$ which is predicted by the first model of the above three models. Then we compute the score of $MS_{1}$ using the other two models and the results are denoted as $S_{1,2}$ and $S_{1,3}$ respectively. The final score of $MS_{1}$ is $S_{1}$ = $S_{1,1}$ + $S_{1,2}$ + $S_{1,3}$. Similarly, $S_{2}$ and $S_{3}$ are calculated. According to biggest score $S_{i}$, the $MS_{i}$ is the optimal sequence predicted by the three models.

\begin{table}
  \caption{F score for each submission in test1}
  \label{tab:ef}
  \begin{tabular}{c|c|c|c}
     \toprule
      & submit1 & submit2 & submit3\\
     \midrule
     task1 & 46.36 & 43.81 & 45.25\\
     task2 & 40.46 & 36.67 & 40.90\\
     \bottomrule
\end{tabular}
\end{table}

\begin{table}
  \caption{F score for each submission in test2 }
  \label{tab:ef}
  \begin{tabular}{c|c|c|c}
     \toprule
     & submit1 & submit2 & submit3\\
     \midrule
     task1 & 46.69 & 43.48 & 45.66\\
     task2 & 35.72 & 32.93 & 34.79\\
     \bottomrule
\end{tabular}
\end{table}

Table 3 and Table 4 show the performance of our each submission. It shows that the Training-22 is not in full accordance with DDI validation guideline and the system proves to be effective as we get the highest F1-score in both two tasks.

\section{Conclusion}
In this paper, we describe our participation in DDI extraction task in TAC 2018. We choose to take part in task1 and task2, which was separately regarded as NER task and RE task from traditional view. Instead, we propose a two-step joint model to recognize mentions as well as relations jointly and build a rule based model to judge the PK subtype. The system proves to be effective as we get the highest F1-score in both two tasks.

\bibliography{acl2018}
\bibliographystyle{acl_natbib}
\end{document}